\documentclass[conference]{IEEEtran}
\IEEEoverridecommandlockouts
\usepackage{cite}
\usepackage{amsmath,amssymb,amsfonts}
\usepackage{algorithmic}
\usepackage{graphicx}
\usepackage{textcomp}
\usepackage{xcolor}
\usepackage{titlesec}
\def\BibTeX{{\rm B\kern-.05em{\sc i\kern-.025em b}\kern-.08em
    T\kern-.1667em\lower.7ex\hbox{E}\kern-.125emX}}
\begin{document}

\title{Reinforcement Learning in Strategy-Based and Atari Games: A Review of Google DeepMind's Innovations\\}

\IEEEoverridecommandlockouts

\author{
    \IEEEauthorblockN{
        \parbox{.45\textwidth}{\centering
            Abdelrhman Shaheen\\
            \textit{Computer Science Engineering Undergraduate Student} \\
            \textit{Egypt Japan University of Science and Technology} \\
            Alexandria, Egypt \\
            abdelrhman.shaheen@ejust.edu.eg
        }
        \hfill
        \parbox{.45\textwidth}{\centering
            Anas Badr\\
            \textit{Computer Science Engineering Undergraduate Student} \\
            \textit{Egypt Japan University of Science and Technology} \\
            Alexandria, Egypt \\
            anas.badr@ejust.edu.eg
        }
    }
    \\[0.5cm]
    \IEEEauthorblockN{
        \parbox{.45\textwidth}{\centering
            Ali Abohendy\\
            \textit{Computer Science Engineering Undergraduate Student} \\
            \textit{Egypt Japan University of Science and Technology} \\
            Alexandria, Egypt \\
            ali.abohendy@ejust.edu.eg
        }
        \hfill
        \parbox{.45\textwidth}{\centering
            Hatem Alsaadawy\\
            \textit{Computer Science Engineering Undergraduate Student} \\
            \textit{Egypt Japan University of Science and Technology} \\
            Alexandria, Egypt \\
            hatem.alsaadawy@ejust.edu.eg
        }
    }
    \\[0.5cm]
    \IEEEauthorblockN{
        \parbox{.45\textwidth}{\centering
            Nadine Alsayad\\
            \textit{Computer Science Engineering Undergraduate Student} \\
            \textit{Egypt Japan University of Science and Technology} \\
            Alexandria, Egypt \\
            nadine.alsayad@ejust.edu.eg
        }
        \hfill
        \parbox{.45\textwidth}{\centering
            Ehab H. El-Shazly\\
            \textit{Department of Computer Science and Engineering, School of Electronics, Communications and Computer Engineering, Egypt-Japan University of Science and Technology (E-JUST), Alexandria, Egypt} \\
            \textit{Egypt Japan University of Science and Technology} \\
            Alexandria, Egypt  \\
            ehab.elshazly@ejust.edu.eg
        }
        \hfill
        \parbox{.45\textwidth}
    }
}

\maketitle
\thispagestyle{plain}
\pagestyle{plain}
\begin{abstract}

    Reinforcement Learning (Rl) has been widely used in many applications, one of
these applications is the field of gaming, which is considered a very good
training ground for AI models. From the innovations of Google DeepMind in this
field using the reinforcement learning algorithms, including model-based,
model-free, and deep Q-network approaches, AlphaGo, AlphaGo Zero, and MuZero.
AlphaGo, the initial model, integrates supervised learning, reinforcement
learning to achieve master in the game of Go, surpassing the performance of
professional human players. AlphaGo Zero refines this approach by eliminating
the dependency on human gameplay data, instead employing self-play to enhance
learning efficiency and model performance. MuZero further extends these
advancements by learning the underlying dynamics of game environments without
explicit knowledge of the rules, achieving adaptability across many games,
including complex Atari games. In this paper, we reviewed the importance of
studying the applications of reinforcement Learning in Atari and strategy-based
games, by discussing these three models, the key innovations of each model,and
how the training process was done; then showing the challenges that every model
faced, how they encounterd them, and how they improved the performance of the
model. We also highlighted the advancements in the field of gaming, including
the advancment in the three models, like the MiniZero and multi-agent models,
showing the future direction for these advancements, and new models from Google
DeepMind.

\end{abstract}


\section{Introduction}
Artificial Intelligence (AI) has revolutionized the gaming industry, both as a
tool for creating intelligent in-game opponents and as a testing environment
for advancing AI research. Games serve as an ideal environment for training and
evaluating AI systems because they provide well-defined rules, measurable
objectives, and diverse challenges. From simple puzzles to complex strategy
games, AI research in gaming has pushed the boundaries of machine learning and
reinforcement learning. Also The benfits from such employment helped game
developers to realize the power of AI methods to analyze large volumes of
player data and optimize game designs. \cite{I1} \\ Atari games, in particular,
with their retro visuals and straightforward mechanics, offer a challenging yet
accessible benchmark for testing AI algorithms. The simplicity of Atari games
hides complexity that they require strategies that involve planning,
adaptability, and fast decision-making, making them also a good testing
environment for evaluating AI's ability to learn and generalize. The
development of AI in games has been a long journey, starting with rule-based
systems and evolving into more sophisticated machine learning models. However,
the machine learning models had a few challenges, from these challenges is that
the games employing AI involves decision making in the game evironment. Machine
learning models are unable to interact with the decisions that the user make
because it depends on learning from datasets and have no interaction with the
environment. To overcome such problem, game developers started to employ
reinforcement learning (RL) in developing games. Years later, deep learning
(DL) was developed and shows remarkable results in video games\cite{I2}. The
combination between both reinforcement learning and deep learning resulted in
Deep Reinforcement Learning (DRL). The first employment of DRL was by game
developers in atari game\cite{I3}. One of the famous companies that employed
DRL in developing AI models is Google DeepMind. This company is known for
developing AI models, including games. Google DeepMind passed through a long
journey in developing AI models for games. Prior to the first DRL game model
they develop, which is AlphaGo, Google DeepMind gave a lot of contributions in
developing DRL, by which these contributions were first employed in Atari
games.\\ For the employment of DRL in games to be efficient, solving tasks in
games need to be sequential, so Google DeepMind combined RL-like techniques
with neural networks to create models capable of learning algorithms and
solving tasks that require memory and reasoning, which is the Neural Turing
Machines (NTMs)\cite{I4}. They then introduced the Deep Q-network (DQN)
algorithm, which is combine deep learning with Q-learning and RL algorithm.
Q-learning is a model in reinforcement learning which use the Q-network, which
is is a type of neural network to approximate the Q-function, which predicts
the value of taking a particular action in a given state\cite{I5}. The DQN
algorithm was the first algorithm that was able to learn directly from
high-dimensional sensory input, the data that have a large number of features
or dimensions\cite{I6}.\\ To enhance the speed of learning in reinforcement
learning agents, Google DeepMind introduced the concept of experience replay,
which is a technique that randomly samples previous experiences from the
agent's memory to break the correlation between experiences and stabilize the
learning process\cite{I7}. They then developed asynchronus methods for DRL,
which is the Actor-Critic (A3C) model. This model showed faster and more stable
training and showed a remarkable performance in Atari games\cite{I8}. By the
usage of these algorithms, Google DeepMind was able to develop the first AI
model that was able to beat the world champion in the game of Go, which is
AlphaGo in 2016.\\
The paper is organized as follows: Section II presents the related work that
surveys the development of DRL in games and the contribution that we 
added to the previous surveys. Section III presents the background 
information about the development of DRL in games. Section IV presents the
first AI model that Google DeepMind developed, which is AlphaGo. Section V
presents AlphaGo Zero. Section VI presents MuZero. Section VII presents the
advancements that were made in developing AI models for games. Section VIII 
presents the future directions that AI models for games will take, and their
applications in real life.
\section{Related Work}
There are a lot of related work that reviewed the
reinforcement learning in strategy-based and atari games. Arulkumaran et
al\cite{I9} this paper serves as a foundational reference that outlines 
the evolution and state-of-the-art developments in DRL up to its 
publication. It also offers insights into how combining deep learning 
with reinforcement learning has led to significant advancements in 
areas such as game playing, robotics, and autonomous decision-making 
systems. Zhao et al.\cite{I10} surveys how DRL combines 
capabilities of deep learning with the decision-making processes of 
reinforcement learning, enabling systems to make control decisions 
directly from input images. It also analysis the development of 
AlphaGo, and examines the algorithms and techniques that contributed 
to AlphaGo's success, providing insights into the integration of DRL 
in complex decision-making tasks. Tang et
al.\cite{I11} also surveys how AlphaGo marked a significant 
milestone by defeating human champions in the game of Go, and its 
architecture and training process; then delves into AlphaGo Zero. 
Shao et al.\cite{I12} categorize DRL methods into three primary 
approaches: value-based, policy gradient, and model-based algorithms, 
offering a comparative analysis of their techniques and properties.
The survey delves into the implementation of DRL across various video 
game types, ranging from classic arcade games to complex real-time 
strategy games.
It highlights how DRL agents, equipped with deep neural network-based 
policies, process high-dimensional inputs to make decisions that 
maximize returns in an end-to-end learning framework.
this review also discusses the achievement of superhuman performance 
by DRL agents in several games, underscoring the significant progress 
in this field.
However, it also addresses ongoing challenges such as exploration-exploitation 
balance, sample efficiency, generalization and transfer learning, 
multi-agent coordination, handling imperfect information, and managing 
delayed sparse rewards.\\
Our paper is similar to Shao et al.\cite{I12}, as we discussed the 
developments that Google DeepMind made in developing AI models for 
games and the advancments that they made over the last years to develop 
the models and the future directions of implementating the DRL in games; 
how this implementation helps in developing real life applications. The
main contribution in our paper is the comprehensive details of the three 
models AlphaGo, AlphaGo Zero, and MuZero, focusing on the key Innovations for 
each model, how the training process was done, challenges that each model
faced and the improvements that were made, and the preformance benchmarks. 
Studying each on of these models in details helps in understanding how 
RL was developed in games reaching to the current state, by which it is 
now used in real life applications.
Also we discussed the advancments in these three AI models, 
reaching to the future directions.

\section{Background}

\renewcommand\theparagraph{\arabic{subsubsection}.\arabic{paragraph}}
\titleformat{\paragraph}[hang]{\normalfont\normalsize\bfseries}{\theparagraph}{1em}{}
\titlespacing*{\paragraph}{2em}{0.5ex plus 0.2ex minus .2ex}{0em}

Reinforcement Learning (RL) is a key area of machine learning that focuses on
learning through interaction with the environment. In RL, an agent takes
actions (A) in specific states (S) with the goal of maximizing the rewards (R)
received from the environment. The foundations of RL can be traced back to
1911, when Thorndike introduced the Law of Effect, suggesting that actions
leading to favorable outcomes are more likely to be repeated, while those
causing discomfort are less likely to recur \cite{bg1}.\\ RL emulates the human
learning process of trial and error. The agent receives positive rewards for
beneficial actions and negative rewards for detrimental ones, enabling it to
refine its policy function—a strategy that dictates the best action to take in
each state. That's said, for a give agent in state $u$, if it takes action $u$,
then the immediate reward $r$ can be modeled as $r(x, u) = \mathbb{E}[r_t \mid
    x=x_{t-1}, u=u_{t-1}]$.\\ So for a full episode of $T$ steps, the cumulative
reward $R$ can be modeled as $R = \sum_{t=1}^{T} r_t$.\\
\subsection{Markov Decision Process (MDP)}

In reinforcement learning, the environment is often modeled as a \textbf{Markov
    Decision Process (MDP)}, which is defined as a tuple $(S, A, P, R, \gamma)$,
where:
\begin{itemize}
    \item \( S \) is the set of states,
    \item \( A \) is the set of actions,
    \item \( P \) is the transition probability function,
    \item \( R \) is the reward function, and
    \item \( \gamma \) is the discount factor.
\end{itemize}

The MDP framework is grounded in \textbf{sequential decision-making}, where the
agent makes decisions at each time step based on its current state. This
process adheres to the \textbf{Markov property}, which asserts that the future
state and reward depend only on the present state and action, not on the
history of past states and actions.

Formally, the Markov property is represented by:

\begin{equation}
    P(s'\mid s, a) = \mathbb{P}[s_{t+1} = s' \mid s_t = s, a_t = a]
\end{equation}

which denotes the probability of transitioning from state $s$ to state $s'$
when action $a$ is taken.

The reward function \( R \) is similarly defined as:

\begin{equation}
    R(s, a) = \mathbb{E}[r_t \mid s_{t-1} = s, a_{t-1} = a]
\end{equation}

which represents the expected reward received after taking action $a$ in state
$S$.
\subsection{Policy and Value Functions}
In reinforcement learning, an agent's goal is to find the optimal policy that
the agent should follow to maximize cumulative rewards over time. To facilitate
this process, we need to quantify the desirability of a given state, which is
done through the \textbf{value function} $V(s)$. Value function estimates the
expected cumulative reward an agent will receive starting from state \( s \)
and continuing thereafter. In essence, the value function reflects how
beneficial it is to be in a particular state, guiding the agent's
decision-making process. The \textbf{state-value function} is then defined as:

\begin{equation}\label{eq:v_pi}
    \begin{split}
        V_\pi(s) & = \mathbb{E_\pi}[G_t \mid s_t = s]                            \\
                 & = \mathbb{E_\pi}[r_t + \gamma r_{t+1}  + \ldots \mid s_t = s]
    \end{split}
\end{equation}

where \( G_t \) is the cumulative reward from time step $t$ onwards. From here
we can define another value function the \textbf{action-value function} under
policy $\pi$, which is $Q_\pi(s, a)$, that estimates the expected cumulative
reward from the state $s$ and taking action $a$ and then following policy
$\pi$:

\begin{equation}\label{eq:q_pi}
    \begin{split}
        Q_\pi(s, a) & = \mathbb{E_\pi}[G_t \mid s_t = s, a_t = a]                            \\
                    & = \mathbb{E_\pi}[r_t + \gamma r_{t+1}  + \ldots \mid s_t = s, a_t = a]
    \end{split}
\end{equation}

where $\gamma$ is the discount factor, which is a decimal value between 0 and 1
that detemines how much we care about immediate rewards versus future reward
rewards \cite{bg2}.\\

We say that a policy $\pi$ is better than another policy $\pi'$ if the expected
return of every state under $\pi$ is greater than or equal to the expected
return of every state under $\pi'$, i.e., $V_\pi(s) \geq V_{\pi'}(s)$ for all
$s \in S$. Eventually, there will be a policy (or policies) that are better
than or equal to all other policies, this is called the \textbf{optimal policy}
$\pi^*$. All optimal policies will then share the same optimal state-value
function $V^*(s)$ and the same optimal action-value function $Q^*(s, a)$, which
are defined as:

\begin{equation}
    \begin{split}
        V^*(s)    & = \max_\pi V_\pi(s)    \\
        Q^*(s, a) & = \max_\pi Q_\pi(s, a)
    \end{split}
\end{equation}

If we can estimate the optimal state-value (or action-value) function, then the
optimal policy $\pi^*$ can be obtained by selecting the action that maximizes
the state-value (or action-value) function at each state, i.e., $\pi^*(s) =
    \arg\max_a Q^*(s, a)$ and that's the goal of reinforcement learning\cite{bg2}.

\subsection{Reinforcement Learning Algorithms}
There are multiple reinforcement learning algorithms that have been developed
that falls under a lot of categories. But, for the sake of this review, we will
focus on the following algorithms that have been used by the Google DeepMind
team in their reinforcement learning models.\\

\subsubsection{\textbf{Model-Based Algorithms: Dynamic Programming}}

\hspace{1em} Dynamic programming (DP) algorithms can be applied when we have a perfect model
of the environment, represented by the transition probability function \( P(s',
r \mid s, a) \). These algorithms rely on solving the Bellman equations
(recursive form of equations \ref{eq:v_pi} and \ref{eq:q_pi}) iteratively to
compute optimal policies. The process alternates between two key steps:
\textbf{policy evaluation} and \textbf{policy improvement}.

\paragraph{Policy Evaluation}
Policy evaluation involves computing the value function \( V^\pi(s) \) under a
given policy \( \pi \). This is achieved iteratively by updating the value of
each state based on the Bellman equation:
\begin{equation}
    V^\pi(s) = \sum_{a \in A} \pi(a \mid s) \sum_{s', r} P(s', r \mid s, a) \left[ r + \gamma V^\pi(s') \right].
\end{equation}

Starting with an arbitrary initial value \( V^\pi(s) \), the updates are
repeated for all states until the value function converges to a stable
estimate.

\paragraph{Policy Improvement}
Once the value function \( V^\pi(s) \) has been computed, the policy is
improved by choosing the action \( a \) that maximizes the expected return for
each state:
\begin{equation}
    \pi'(s) = \arg\max_a \sum_{s', r} P(s', r \mid s, a) \left[ r + \gamma V^\pi(s') \right].
\end{equation}

This step ensures that the updated policy \( \pi' \) is better than or equal to
the previous policy \( \pi \). The process alternates between policy evaluation
and improvement until the policy converges to the optimal policy \( \pi^* \),
where no further improvement is possible. It can be visualized as:

\[
    \pi_0 \xrightarrow{\text{Eval}} V^{\pi_0} \xrightarrow{\text{Improve}} \pi_1 \xrightarrow{\text{Eval}} V^{\pi_1} \xrightarrow{\text{Improve}} \pi_2 \xrightarrow{\text{Eval}} \ldots \xrightarrow{\text{Improve}} \pi^*.
\]

\paragraph{Value Iteration}
Value iteration simplifies the DP process by combining policy evaluation and
policy improvement into a single step. Instead of evaluating a policy
completely, it directly updates the value function using:
\begin{equation}
    V^*(s) = \max_a \sum_{s', r} P(s', r \mid s, a) \left[ r + \gamma V^*(s') \right].
\end{equation}

This method iteratively updates the value of each state until convergence and
implicitly determines the optimal policy. Then the optimal policy can be
obtained by selecting the action that maximizes the value function at each
state, as
\begin{equation}
    \pi^*(s) = \arg\max_a \sum_{s', r} P(s', r \mid s, a) \left[ r + \gamma V^*(s') \right].
\end{equation}

Dynamic Programming's systematic approach to policy evaluation and improvement
forms the foundation for the techniques that have been cruical in training
systems like AlphaGo Zero and MuZero.

\subsubsection{\textbf{Model-Free Algorithms}}

\paragraph{Monte Carlo Algorithm (MC)}
The Monte Carlo (MC) algorithm is a model-free reinforcement learning method
that estimates the value of states or state-action pairs under a given policy
by averaging the returns of multiple episodes. Unlike DP, MC does not require a
perfect model of the environment and instead learns from sampled experiences.

The key steps in MC include:

\begin{itemize}
    \item \textbf{Policy Evaluation:} Estimate the value of a state or state-action pair \( Q(s, a) \) by averaging the returns observed in multiple episodes.
    \item \textbf{Policy Improvement:} Update the policy \( \pi \) to choose actions that maximize the estimated value \( Q(s, a) \).
\end{itemize}

MC algorithms operate on complete episodes, requiring the agent to explore all
state-action pairs sufficiently to ensure accurate value estimates. The updated
policy is given by:
\begin{equation}
    \pi(s) = \arg\max_a Q(s, a).
\end{equation}

While both MC and DP alternate between policy evaluation and policy
improvement, MC works with sampled data, making it suitable for environments
where the dynamics are unknown or difficult to model.

This algorithm is particularly well-suited for environments that are
\emph{episodic}, where each episode ends in a terminal state after a finite
number of steps. \\ Monte Carlo's reliance on episodic sampling and policy
refinement has directly influenced the development of search-based methods like
Monte Carlo Tree Search (MCTS), which was crucial in AlphaGo for evaluating
potential moves during gameplay. The algorithm's adaptability to model-free
settings has made it a cornerstone of modern reinforcement learning strategies.

\paragraph{Temporal Diffrence (TD)}

Temporal Diffrence is another model free algorithm that's very similar To Monte
Carlo, but instead of waiting for termination of the episode to give the
return, it estimates the return based on the next state. The key idea behind TD
is to update the value function based on the difference between the current
estimate and the estimate of the next state. The TD return is then given by:
\begin{equation}
    G_t = r_{t+1} + \gamma V(s_{t+1})
\end{equation}
that's the target (return value estimation) of state $s$ at time $t$ is the immediate reward $r$ plus the
discounted value of the next state $s_{t+1}$.\\
This here is called the TD(0) algorithm, which is the simplest form of TD
that take only one future step into account. The update rule for TD(0) is:
\begin{equation}
    V(s_t) = V(s_t) + \alpha [r_{t+1} + \gamma V(s_{t+1}) - V(s_t)]
\end{equation}

There are other temporal difference algorithms that works exactly like TD(0),
but with more future steps, like TD($\lambda$). \\

Another important variant of TD is the Q-learning algorithm, which is an
off-policy TD algorithm that estimates the optimal action-value function $Q^*$
by updating the current action value based on the optimal action value of the
next state. The update rule for Q-learning is:

\begin{equation}
    Q(s_t, a_t) = Q(s_t, a_t) + \alpha[r_{t+1} + \gamma \max_a Q(s_{t+1}, a) - Q(s_t, a_t)] .
\end{equation}

and after the algorithm converges, the optimal policy can be obtained by
selecting the action that maximizes the action-value function at each state, as
$\pi^*(s) = \arg\max_a Q^*(s, a)$.

Temporal Difference methods, including Q-learning, play a crucial role in
modern reinforcement learning by enabling model-free value function estimation
and action selection without the need to terminate the episode. In systems like
AlphaGo and MuZero, TD methods are used to update value functions efficiently
and support complex decision-making processes without requiring a model of the
environment.

\subsubsection{\textbf{Deep RL: Deep Q-Network (DQN)}}

Deep Q-Networks (DQN) represent a significant leap forward in the integration
of deep learning with reinforcement learning. DQN extends the traditional
Q-learning algorithm by using a deep neural network to approximate the Q-value
function, which is essential in environments with large state spaces where
traditional tabular methods like Q-learning become infeasible.

In standard Q-learning, the action-value function \(Q(s, a)\) is learned
iteratively based on the Bellman equation, which updates the Q-values according
to the reward received and the value of the next state. However, when dealing
with complex, high-dimensional inputs such as images or unstructured data, a
direct tabular representation of the Q-values is not practical. This is where
DQN comes in: it uses a neural network, typically a convolutional neural
network (CNN), to approximate \(Q(s, a; \theta)\), where \(\theta\) represents
the parameters of the network.

The core ideas behind DQN are similar to those of traditional Q-learning, but
with a few key innovations that address issues such as instability and high
variance during training. The DQN algorithm introduces the following
components:

\begin{itemize}
    \item \textbf{Experience Replay:} To improve the stability of training and to break the correlation between consecutive experiences, DQN stores the agent’s experiences in a replay buffer. Mini-batches of experiences are randomly sampled from this buffer to update the network, which helps in better generalization.
    \item \textbf{Target Network:} DQN uses two networks: the primary Q-network and a target Q-network. The target network is updated less frequently than the primary network and is used to calculate the target value in the Bellman update. This reduces the risk of oscillations and divergence during training.
\end{itemize}

The update rule for DQN is based on the Bellman equation for Q-learning, but
with the neural network approximation:

\begin{equation}
    \begin{split}
         & Q(s_t, a_t; \theta) = Q(s_t, a_t; \theta) +                                                     \\
         & \alpha \left[ r_{t+1} + \gamma \max_{a'} Q(s_{t+1}, a'; \theta^-) - Q(s_t, a_t; \theta) \right]
    \end{split}
\end{equation}

where \( \theta^- \) represents the parameters of the target network. By
training the network to minimize the difference between the predicted Q-values
and the target Q-values, the agent learns an optimal policy over
time.\cite{bg4}

The DQN algorithm revolutionized reinforcement learning, especially in
applications requiring decision-making in high-dimensional spaces. One of the
most notable achievements of DQN was its success in mastering a variety of
Atari 2600 games directly from raw pixel input, achieving human-level
performance across multiple games. This breakthrough demonstrated the power of
combining deep learning with reinforcement learning to solve complex,
high-dimensional problems.

In subsequent improvements, such as Double DQN, Dueling DQN, and Prioritized
Experience Replay, enhancements were made to further stabilize training and
improve performance. However, the foundational concepts of using deep neural
networks to approximate Q-values and leveraging experience replay and target
networks remain core to the DQN framework.

\section{AlphaGo}
\subsection{Introduction}
AlphaGo is a groundbreaking AI model that utilizes neural networks and tree
search to play the game of Go, which is thought to be one of the most
challenging classic games for artificial intelligence owing to its enormous
search space and the difficulty of evaluating board positions and moves
\cite{Silver2016}. \\\\ AlphaGo uses value networks for position evaluation and
policy networks for taking actions, that combined with Monte Carlo simulation
achieved a 99.8\% winning rate, and beating the European human Go champion in 5
out 5 games.

\subsection{Key Innovations}

\subsubsection*{Integration of Policy and Value Networks with MCTS}
AlphaGo combines policy and value networks in an MCTS framework to efficiently explore and evaluate the game tree. Each edge \( (s, a) \) in the search tree stores:
\begin{itemize}
    \item Action value \( Q(s, a) \): The average reward for taking action \( a \) from
          state \( s \).
    \item Visit count \( N(s, a) \): The number of times this action has been explored.
    \item Prior probability \( P(s, a) \): The probability of selecting action \( a \),
          provided by the policy network.
\end{itemize}

During the selection phase, actions are chosen to maximize:
\begin{equation}
    a_t = \arg\max_a \left( Q(s, a) + u(s, a) \right)
\end{equation}

where the exploration bonus \( u(s, a) \) is defined as:
\begin{equation}
    u(s, a) \propto \frac{P(s, a)}{1 + N(s, a)}
\end{equation}

When a simulation reaches a leaf node, its value is evaluated in two ways: 1.
Value Network Evaluation: A forward pass through the value network predicts \(
v_\theta(s) \), the likelihood of winning. 2. Rollout Evaluation: A lightweight
policy simulates the game to its conclusion, and the terminal result \( z \) is
recorded.

These evaluations are combined using a mixing parameter \( \lambda \):
\begin{equation}
    V(s_L) = \lambda v_\theta(s_L) + (1 - \lambda) z_L
\end{equation}

The back propagation step updates the statistics of all nodes along the path
from the root to the leaf. \\\\ It's also worth noting that the SL policy
network performed better than the RL policy network and that's probably because
humans select a diverse beam of promising moves, whereas RL optimizes for the
single best move.

Conversely though, the value network that was derived from the RL policy
performed better than the one derived from the SL policy.
\begin{figure}[htbp]
    \centering
    \includegraphics[width=0.9\linewidth, keepaspectratio]{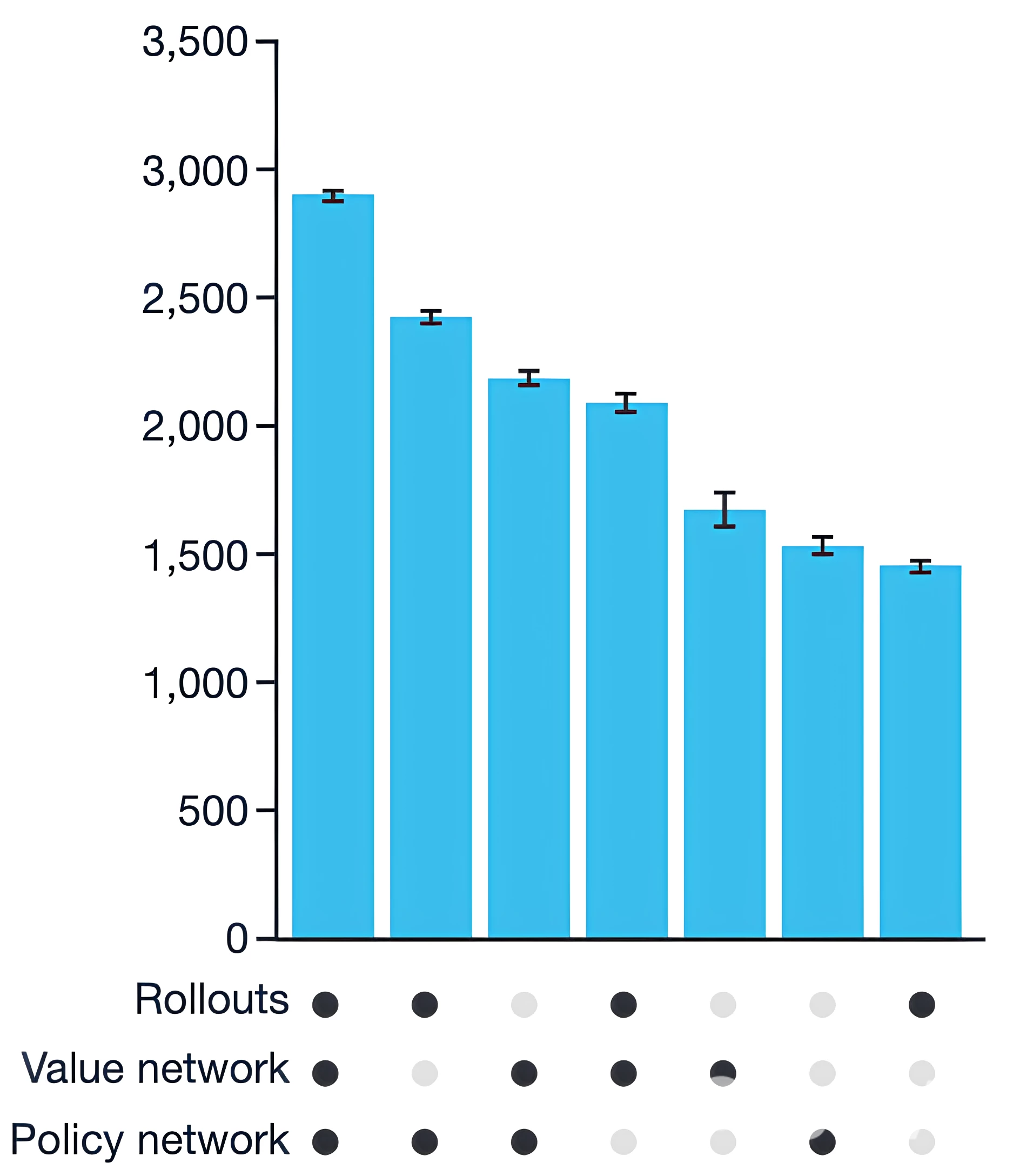}
    \caption{Performance of AlphaGo, on a single machine, for different
combinations of components.}
\end{figure}

\subsection{Training Process}

\subsubsection{Supervised Learning for Policy Networks}
The policy network was initially trained using supervised learning on human
expert games. The training data consisted of 30 million board positions sampled
from professional games on the KGS Go Server. The goal was to maximize the
likelihood of selecting the human move for each position:
\begin{equation}
    \Delta \sigma \propto \nabla_\sigma \log p_\sigma(a | s)
\end{equation}

where \( p_\sigma(a | s) \) is the probability of selecting action \( a \)
given state \( s \).

This supervised learning approach achieved a move prediction accuracy of 57.0\%
on the test set, significantly outperforming prior methods. This stage provided
a solid foundation for replicating human expertise.

\subsubsection{Reinforcement Learning for Policy Networks}
The supervised learning network was further refined through reinforcement
learning (RL). The weights of the RL policy network were initialized from the
SL network. AlphaGo then engaged in self-play, where the RL policy network
played against earlier versions of itself to iteratively improve.

The reward function used for RL was defined as:
\begin{equation}
    r(s) =
    \begin{cases}
        +1 & \text{if win}                           \\
        -1 & \text{if loss}                          \\
        0  & \text{otherwise (non-terminal states).}
    \end{cases}
\end{equation}

At each time step \( t \), the network updated its weights to maximize the
expected reward using the policy gradient method:
\begin{equation}
    \Delta \rho \propto z_t \nabla_\rho \log p_\rho(a_t | s_t)
\end{equation}

where \( z_t \) is the final game outcome from the perspective of the current
player.

This self-play strategy allowed AlphaGo to discover novel strategies beyond
human knowledge. The RL policy network outperformed the SL network with an 80\%
win rate and achieved an 85\% win rate against Pachi, a strong open-source Go
program, without using MCTS.

\subsubsection{Value Network Training}
The value network was designed to evaluate board positions by predicting the
likelihood of winning from a given state. Unlike the policy network, it outputs
a single scalar value \( v_\theta(s) \) between \(-1\) (loss) and \(+1\) (win).

Training the value network on full games led to overfitting due to the strong
correlation between successive positions in the same game. To mitigate this, a
new dataset of 30 million distinct board positions was generated through
self-play, ensuring that positions came from diverse contexts.

The value network was trained by minimizing the mean squared error (MSE)
between its predictions \( v_\theta(s) \) and the actual game outcomes \( z \):
\begin{equation}
    L(\theta) = \mathbb{E}_{(s, z) \sim D} \left[ (v_\theta(s) - z)^2 \right]
\end{equation}

\begin{figure}[htbp]
    \centering
    \includegraphics[width=\linewidth, keepaspectratio]{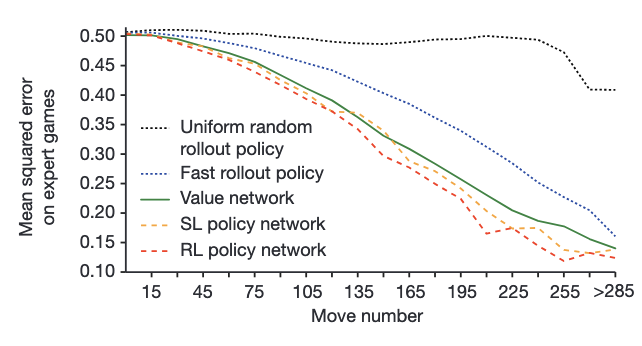}
    \caption{Comparison of evaluation
accuracy between the value network and rollouts with different policies.}
\end{figure}

\subsection{Challenges and Solutions}
AlphaGo overcame several challenges:
\begin{itemize}
    \item Overfitting: Training the value network on full games led to memorization. This
          was mitigated by generating a diverse self-play dataset.
    \item Scalability: Combining neural networks with MCTS required significant
          computational resources, addressed through parallel processing on GPUs and
          CPUs.
    \item Exploration vs. Exploitation: Balancing these in MCTS was achieved using the
          exploration bonus \( u(s, a) \) and the policy network priors.
\end{itemize}

\subsection{Performance Benchmarks}
AlphaGo achieved the following milestones:
\begin{itemize}
    \item 85\% win rate against Pachi without using MCTS.
    \item 99.8\% win rate against other Go programs in a tournament held to evaluate the performance of AlphaGo.
    \item Won 77\%, 86\%, and 99\% of handicap games against Crazy Stone, Zen and Pachi,
          respectively.
    \item Victory against professional human players such as Fan Hui (5-0) and Lee Sedol
          (4-1), marking a significant breakthrough in AI.
\end{itemize}

\begin{figure}[htbp]
    \centering
    \includegraphics[width=\linewidth, keepaspectratio]{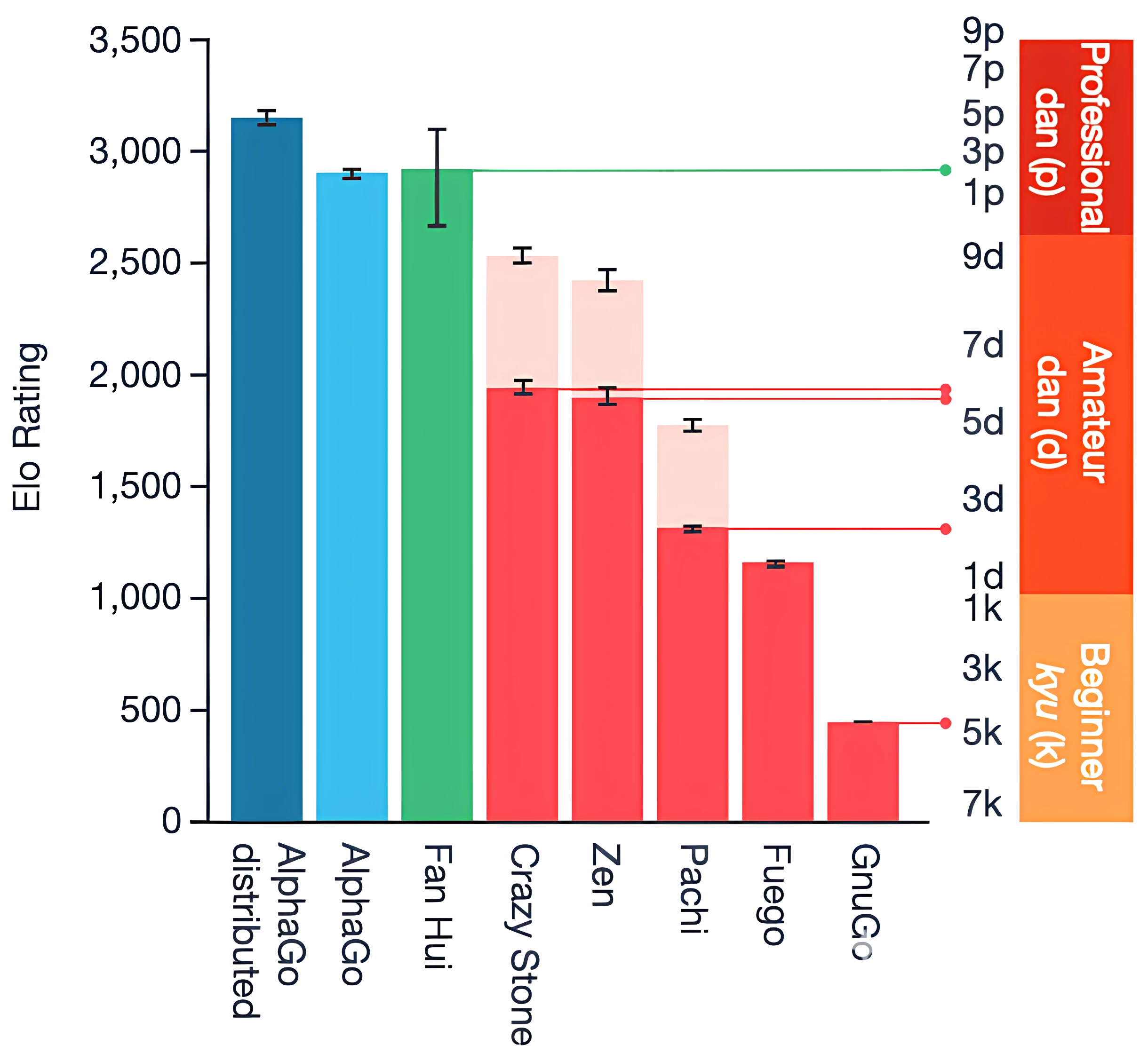}
    \caption{Elo rating comparison between AlphaGo and other Go programs.}
\end{figure}

\section{AlphaGo Zero}
\subsection{Introduction}
AlphaGo Zero represents a groundbreaking advancement in artificial intelligence
and reinforcement learning. Unlike its predecessor, AlphaGo, which relied on
human gameplay data for training, AlphaGo Zero learns entirely from self-play,
employing deep neural networks and Monte Carlo Tree Search (MCTS). \cite{agz1}
\\\\ By starting with only the rules of the game and leveraging reinforcement
learning, AlphaGo Zero achieved superhuman performance in Go, defeating the
previous version of AlphaGo in a 100-0 match.
\subsection{Key Innovations}

AlphaGo Zero introduced several groundbreaking advancements over its
predecessor, AlphaGo, streamlining and enhancing its architecture and training
process:

\begin{enumerate}
    \item Unified Neural Network \( f_\theta \): AlphaGo Zero replaced AlphaGo's
          dual-network setup—separate networks for policy and value—with a single neural
          network \( f_\theta \). This network outputs both the policy vector \( p \) and
          the value scalar \( v \) for a given game state, reprsented as

          \begin{equation}
              f_\theta(s) = (p, v)
          \end{equation}

          This unified architecture simplifies the model and improves training
          efficiency.

    \item Self-Play Training: Unlike AlphaGo, which relied on human games as training
          data, AlphaGo Zero was trained entirely through self-play. Starting from random
          moves, it learned by iteratively playing against itself, generating data and
          refining \( f_\theta \) without any prior knowledge of human strategies. This
          removed biases inherent in human gameplay and allowed AlphaGo Zero to discover
          novel and highly effective strategies.

    \item Removal of Rollouts: AlphaGo Zero eliminated the need for rollouts, which were
          computationally expensive simulations to the end of the game used by AlphaGo's
          MCTS. Instead, \( f_\theta \) directly predicts the value \( v \) of a state,
          providing a more efficient and accurate estimation.

    \item Superior Performance: By integrating these advancements, AlphaGo Zero defeated
          AlphaGo 100-0 in direct matches, demonstrating the superiority of its self-play
          training, unified architecture, and reliance on raw rules over pre-trained
          human data.
\end{enumerate}

\subsection{Training Process}

\subsubsection{Monte Carlo Tree Search (MCTS) as policy evaluation operator}
Intially the neural network \( f_\theta \) is not very accurate in predicting
the best move, as it is intiallised with random weights at first. To overcome
this, AlphaGo Zero uses MCTS to explore the game tree and improve the policy.
\\\\ At a given state S, MCTS expands simualtions of the best moved that are
most likely to generate a win based on the initial policy $P(s,a)$ and the
value $V$. MCTS iteratively selects moves that maximize the upper confidence
bound (UCB) of the action value. UCB is designed to balanced exploration and
exploitation. and it is defined as

\begin{equation}
    UCB = Q(s, a) + U(s, a)
\end{equation}

where \[U(s, a) \propto \frac{p(s, a)}{1 + N(s, a)}\] MCTS at the end of the search returns the policy vector \( \pi \) which is used
to update the neural network \( f_\theta \) by minimizing the cross-entropy
loss between the predicted policy by $f_\theta$ and the MCTS policy.

\subsubsection{Policy Iteration and self play}
The agent plays games against itself using the predicted policy $P(s, a)$. The
agent uses the MCTS to select the best move at each state and the game is
played till the end in a process called self play. The agent then uses the
outcome of the game, $z$ game winner and $\pi$ to update the neural network.
This process is repeated for a large number of iterations.

\subsubsection{Network Training Process}

The neural network is updated after each self-play game by using the data
collected during the game. This process involves several key steps:
\begin{enumerate}
    \item Intilisation of the network: The neural network starts with random weights
          $\theta_0$, as there is no prior knowledge about the game.
    \item Generating Self-play Games: For each iteration $i \geq 1$ self-play games are
          generated. During the game, the neural network uses its current parameters
          $\theta_{i - 1}$ to run MCTS and generate search probabilities $\pi_t$ for each
          move at time step $t$.
    \item Game Termination and scoring: A game ends when either both players pass, a
          resignation threshold is met, or the game exceeds a maximum length. The winner
          of the game is determined, and the final result $z_t$ is recorded, providing
          feedback to the model.
    \item Data Colletion: for each time step $t$, the training data $(s_t, \pi_t, z_t)$
          is stored, where $s_t$ is the game state, $\pi_t$ is the search probabilities,
          and $z_t$ is the game outcome.
    \item Network training process: after collecing data from self-play, The neural
          network $f_\theta$ is adjusted to minimize the error between the predicted
          value v and the self-play winnder z, and to maximize the similarity between the
          search probabilities $P$ and the MCTS probabilities. This is done by using a
          loss function that combines the mean-squared error and the cross entropy losses
          repsectibly. The loss function is defined as

          \begin{equation}
              L = (z - v)^2 - \pi^T \log p + c||\theta||^2
          \end{equation}

          where $c$ is the L2 regularization term.

\end{enumerate}

\subsection{Challenges and Solutions}
Alpha Go Zero overcame several challanges:
\begin{enumerate}
    \item Human knowledge Dependency: AlphaGo Zero eliminated the need for human gameplay
          data, relying solely on self-play to learn the game of Go. This allowed it to
          discover novel strategies that surpassed human expertise.
    \item Compelxity of the dual network approach in alpha go: AlphaGo utilized separate
          neural networks for policy prediction $p$ and value estimation $V$, increasing
          the computational complexity. AlphaGo Zero unified these into a \textbf{single
              network} that outputs both $p$ and $V$, simplifying the architecture and
          improving training efficiency.
    \item The need of handcrafted features: AlphaGo relied on handcrafted features, such
          as board symmetry and pre-defined game heuristics, for feature extraction.
          AlphaGo Zero eliminated the need for feature engineering by using \textbf{raw
              board states} as input, learning representations directly from the data.
\end{enumerate}

\subsection{Performance Benchmarks}

\begin{figure}[ht]
    \centering
    \includegraphics[width=0.35\textwidth]{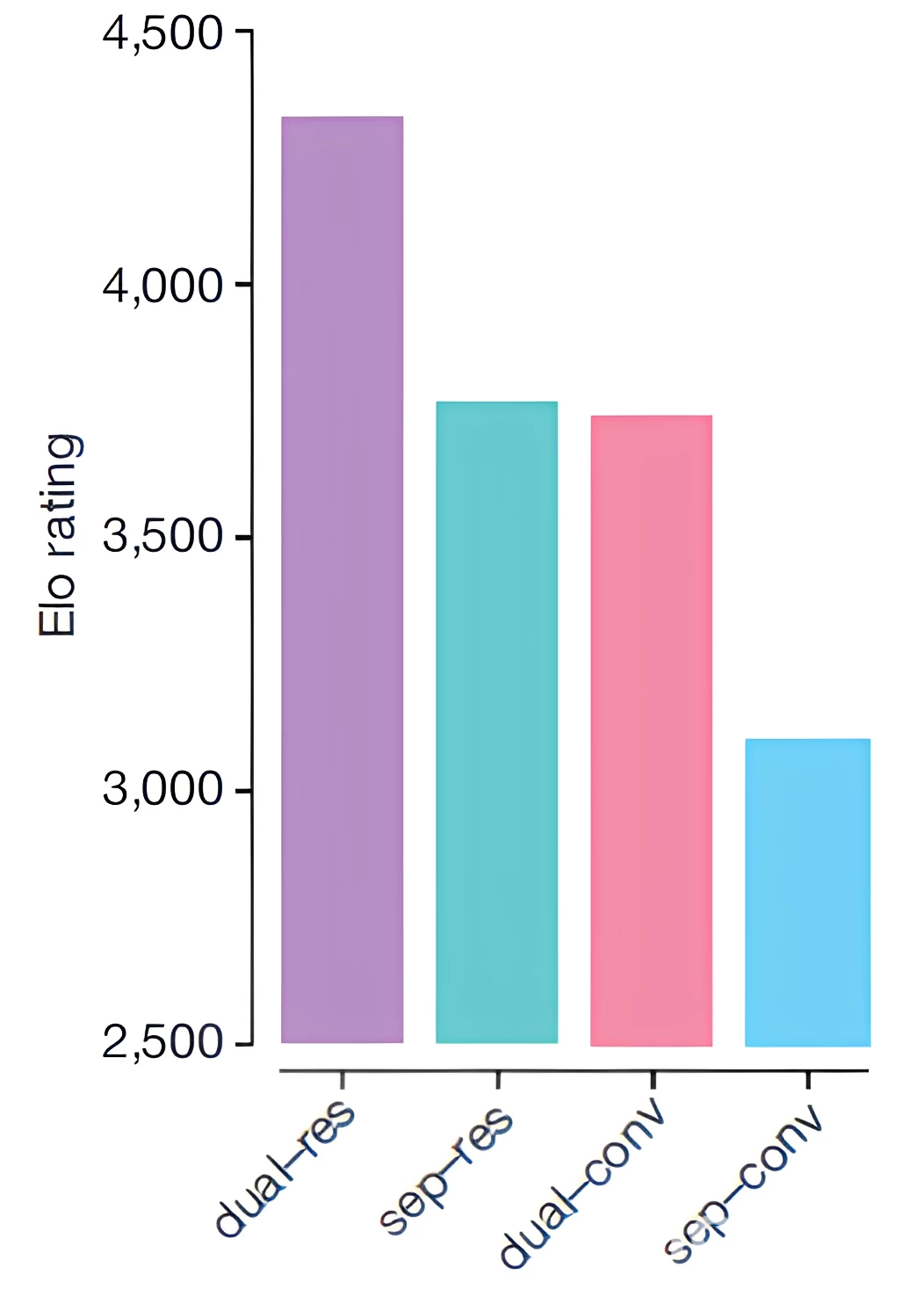}
    \caption{Elo rating comparison of different neural network architectures.}
    \label{fig:results_agz}
\end{figure}

AlphaGo Zero introduced a significant improvement in neural network
architecture by employing a unified residual network (ResNet) design for its
$f_\theta$ model. This replaced the separate CNN-based architectures previously
used in AlphaGo, which consisted of distinct networks for policy prediction and
value estimation. \\\\ The superiority of this approach is evident in the Elo
rating comparison shown in fig.\ref{fig:results_agz}. The "dual-res"
architecture, utilized in AlphaGo Zero, achieved the highest Elo rating,
significantly outperforming other architectures like "dual-conv" and "sep-conv"
used in earlier versions of AlphaGo.

\section{MuZero}
\subsection{Introduction}
Through the development of AlphaZero, a general model for board games with
superhuman ability has been achieved in three games: Go, chess, and Shogi. It
could achieve these results without the need for human gameplay data or
history, instead using self-play in an enclosed environment. However, the model
still relied on a simulator that could perfectly replicate the behavior, which
might not translate well to real-world applications, where modeling the system
might not be feasible. MuZero was developed to address this challenge by
developing a model-based RL approach that could learn without explicitly
modeling the real environment. This allowed for the same general approach used
in AlphaZero to be used in Atari environments where reconstructing the
environment is costly. Essentially, MuZero was deployed to all the games with
no prior knowledge of them or specific optimization and managed to show
state-of-the-art results in almost all of them.

\subsection{MuZero Algorithm}
\begin{figure*}[t]
    \centering
    \includegraphics[width=0.8\textwidth]{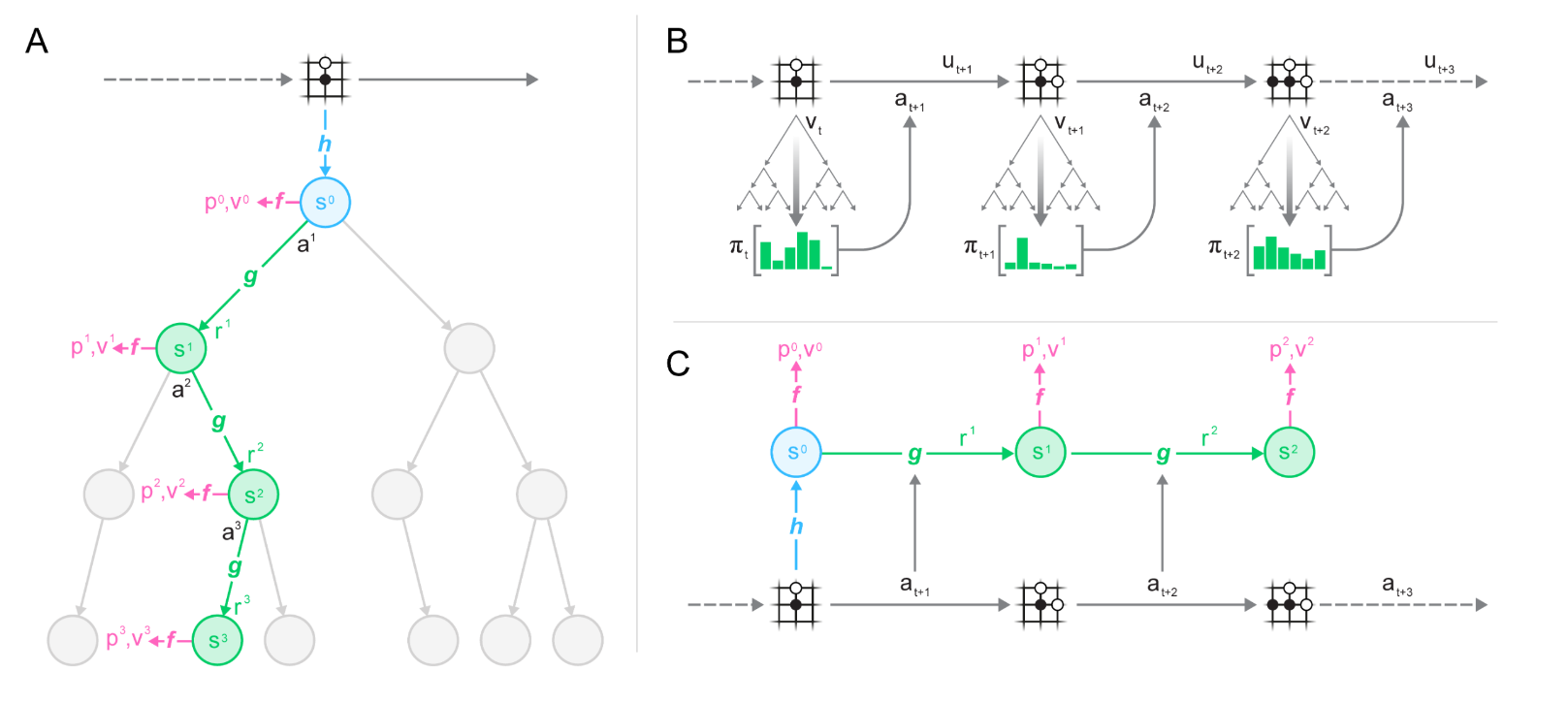}
    \caption{(A) Represents the progression of the model through its MDP, while (B) Represents MuZero acting as an environment with MCTS as feedback, and (C) Represents a diagram of training MuZero's model.}
\end{figure*}
The model takes in an input of observations $o_1, \ldots, o_t$ that are then fed to a representation network $h$,
which reduces the dimensions of the input and produces a root hidden state $s_0$. Internally, the model
mirrors an MDP, with each state representing a node with edges connecting it to the future states
depending on available actions. Unlike traditional RL approaches, this hidden state is not constrained to
contain the information necessary to reproduce the entire future observations. Instead, the hidden states
are only optimized for predicting information that is related to planning. At every time step, the model
predicts the policy, the immediate reward, and the value function. The output of the state-action pair is
then used by the dynamics function to produce future states. Similar to AlphaZero, a Monte Carlo tree
search is used to find the best action policy given an input space. This is used to train the model by
comparing the MCTS policy with the predictor function policy. Also, after a few training runs, the model
ceases to use illegal moves, and the predicted actions map to the real action space. This eliminates the
need for a simulator, as the model internalizes the environment characteristics it deems necessary for
planning and acting, which generally converges to reality through training. The value function at
the final step is compared against the game result in board games, i.e., win, loss, or a draw.

\subsection{Loss function and learning equations}

\begin{align}
    s_0      & = h_\theta(o_1, \ldots, o_t) \\
    r_k, s_k & = g_\theta(s_{k-1}, a_k)     \\
    p_k, v_k & = f_\theta(s_k)
\end{align}

\begin{equation}
    \setlength{\arraycolsep}{1.5pt} 
    \begin{bmatrix}
        p_k \\ v_k \\ r_k
    \end{bmatrix}
    =
    \mu_\theta(o_1, \ldots, o_t, a_1, \ldots, a_k)
    \label{eq:matrix}
\end{equation}

\begin{align}
    \nu_t, \pi_t & = \text{MCTS}(s_0^t, \mu_\theta) \\
    a_t          & \sim \pi_t
\end{align}

\begin{align}
    p_k^t, v_k^t, r_k^t & = \mu_\theta(o_1, \ldots, o_t, a_{t+1}, \ldots, a_{t+k}) \\
    z_t                 & =
    \begin{cases}
        u_T,                                               & \text{for games}        \\
        u_{t+1} + \gamma u_{t+2} + \ldots \nonumber                                  \\
        \quad + \gamma^{n-1} u_{t+n} + \gamma^n \nu_{t+n}, & \text{for general MDPs}
    \end{cases}
\end{align}

\begin{align}
    l_t(\theta) & =
    \sum_{k=0}^K \big[ l_r(u_{t+k}, r_k^t) + l_v(z_{t+k}, v_k^t) \nonumber \\
                & \quad + l_p(\pi_{t+k}, p_k^t) \big] + c \|\theta\|^2
\end{align}

\begin{align}
    l_r(u, r)   & =
    \begin{cases}
        0,                & \text{for games}        \\
        \phi(u)^T \log r, & \text{for general MDPs}
    \end{cases} \\
    l_v(z, q)   & =
    \begin{cases}
        (z - q)^2,        & \text{for games}        \\
        \phi(z)^T \log q, & \text{for general MDPs}
    \end{cases} \\
    l_p(\pi, p) & = \pi^T \log p
\end{align}

\subsection{MCTS}
MuZero uses the same approach developed in AlphaZero to find the optimum action
given an internal state. MCTS is used where states are the nodes, and the edges
store visit count, mean value, policy, and reward. The search is done in a
three-phase setup: selection, expansion, and backup. The simulation starts with
a root state, and an action is chosen based on the state-transition reward
table. Then, after the end of the tree, a new node is created using the output
of the dynamics function as a value, and the data from the prediction function
is stored in the edge connecting it to the previous state. Finally, the
simulation ends, and the updated trajectory is added to the state-transition
reward table. In two-player zero-sum games, board games, for example, the value
function is bounded between 0 and 1, which is helpful to use value estimation
and probability using the pUCT rule. However, many other environments have
unbounded values, so MuZero rescales the value to the maximum value observed by
the model up to this training step, ensuring no environment-specific data is
needed.\cite{mz1}

\subsection{Results}
The MuZero model demonstrated significant improvements across various test
cases, achieving state-of-the-art performance in several scenarios. Key
findings include:

\subsubsection{Board Games}
\begin{itemize}
    \item When tested on the three board games AlphaZero was trained for (Go, chess, and
          shogi):
          \begin{itemize}
              \item MuZero matched AlphaZero's performance \textbf{without any prior knowledge} of
                    the games' rules.
              \item It achieved this with \textbf{reduced computational cost} due to fewer residual
                    blocks in the representation function.
          \end{itemize}
\end{itemize}

\subsubsection{Atari Games}
\begin{itemize}
    \item MuZero was tested on 60 Atari games, competing against both human players and
          state-of-the-art models (model-based and model-free). Results showed:
          \begin{itemize}
              \item \textbf{Starting from regular positions:} MuZero outperformed competitors in \textbf{46 out of 60 games}.
              \item \textbf{Starting from random positions:} MuZero maintained its lead in \textbf{37 out of 60 games}, though its performance was reduced.
          \end{itemize}
    \item The computational efficiency and generalization of MuZero highlight its
          effectiveness in complex, unstructured environments.
\end{itemize}

\subsubsection{Limitations}
\begin{itemize}
    \item Despite its strengths, MuZero struggled in certain games, such as:
          \begin{itemize}
              \item \emph{Montezuma's Revenge} and \emph{Pitfall}, which require long-term planning and strategy.
          \end{itemize}
    \item General challenges:
          \begin{itemize}
              \item Long-term dependencies remain difficult for MuZero, as is the case for RL
                    models in general.
              \item Limited input space and lack of combinatorial inputs in Atari games could
                    introduce scalability issues for broader applications.\cite{mz1}
          \end{itemize}
\end{itemize}

\section{Advancements}
The evolution of AI in gaming, particularly through the development of AlphaGo,
AlphaGo Zero, and MuZero, highlights remarkable advancements in reinforcement
learning and artificial intelligence. AlphaGo, the pioneering model, combined
supervised learning and reinforcement learning to master the complex game of
Go, setting the stage for AI to exceed human capabilities in well-defined
strategic games. Building on, AlphaGo Zero eliminated the reliance on human
data, introducing a fully self-supervised approach that demonstrated greater
efficiency and performance by learning solely through self-play. MuZero took
this innovation further by generalizing beyond specific games like Go, Chess,
and Shogi, employing model-based reinforcement learning to predict dynamics
without explicitly knowing the rules of the environment. Completing on these
three models, here are some of the advancements that developed from them:
AlphaZero and MiniZero; and one of the most used in generating AI models,
Multi-agent models.
\subsection*{AlphaZero}
While AlphaGo Zero was an impressive feat, designed specifically to master the
ancient game of Go through self-play, AlphaZero developes it by generalizing
its learning framework to include multiple complex games: chess, shogi
(Japanese chess), and Go. The key advancement is in its ability to apply the
same algorithm across different games without requiring game-specific
adjustments. AlphaZero's neural network is trained through self-play,
predicting the move probabilities and game outcomes for various positions. This
prediction is then used to guide the MCTS, which explores potential future
moves and outcomes to determine the best action. Through iterative self-play
and continuous refinement of the neural network, AlphaZero efficiently learns
and improves its strategies across different games\cite{AD3}. Another
significant improvement is in AlphaZero’s generalized algorithm, is that it
does not need to be fine-tuned for each specific game. This was a departure
from AlphaGo Zero’s Go-specific architecture, making AlphaZero a more versatile
AI system.\\ AlphaZero's architecture integrates a single neural network that
evaluates both the best moves and the likelihood of winning from any given
position, streamlining the learning process by eliminating the need for
separate policy and value networks used in earlier systems. This innovation not
only enhances computational efficiency but also enables AlphaZero to adopt
unconventional and creative playing styles that diverge from established human
strategies.
\subsection*{MiniZero}
MiniZero is a a zero-knowledge learning framework that supports four
state-of-the-art algorithms, including AlphaZero, MuZero, Gumbel AlphaZero, and
Gumbel MuZero\cite{AD1}. Gumbel AlphaZero and Gumbel MuZero are variants of the
AlphaZero and MuZero algorithms that incorporate Gumbel noise into their
decision-making process to improve exploration and planning efficiency in
reinforcement learning tasks. Gumbel noise is a type of stochastic noise
sampled from the Gumbel distribution, commonly used in decision-making and
optimization problems.\\ MiniZero is a simplified version of the original
MuZero algorithm, which is designed to be have a more simplified architecture
reducing the complexity of the neural network used to model environment
dynamics, making it easier to implement and experiment with. This
simplification allows MiniZero to perform well in smaller environments with
fewer states and actions, offering faster training times and requiring fewer
computational power compared to MuZero.

\subsection*{Multi-agent models}
Multi-agent models in reinforcement learning (MARL) represent an extension of
traditional single-agent reinforcement learning. In these models, multiple
agents are simultaneously interacting, either competitively or cooperatively,
making decisions that impact both their own outcomes and those of other agents.
The complexity in multi-agent systems arises from the dynamic nature of the
environment, where the actions of each agent can alter the environment and the
states of other agents. Unlike in single-agent environments, where the agent
learns by interacting with a static world, multi-agent systems require agents
to learn not only from their direct experiences but also from the behaviors of
other agents, leading to a more complex learning process. Agents must adapt
their strategies based on what they perceive other agents are doing, and this
leads to problems such as strategic coordination, deception, negotiation, and
competitive dynamics. In competitive scenarios, agents might attempt to outwit
one another, while in cooperative scenarios, they must synchronize their
actions to achieve a common goal\cite{AD2}.\\ AlphaGo and AlphaGo Zero are not
designed to handle multi-agent environments. The core reason lies in their
foundational design, which assumes a single agent interacting with a static
environment. AlphaGo and AlphaGo Zero both rely on model-based reinforcement
learning and self-play, where a single agent learns by interacting with itself
or a fixed opponent, refining its strategy over time. However, these models are
not built to adapt to the dynamic nature of multi-agent environments, where the
state of the world constantly changes due to the actions of other agents. In
AlphaGo and AlphaGo Zero, the environment is well-defined, and the agent’s
objective is to optimize its moves based on a fixed set of rules. The agents in
these models do not need to account for the actions of other agents in
real-time or consider competing strategies, which are essential in multi-agent
systems. Additionally, AlphaGo and AlphaGo Zero are not designed to handle
cooperation or negotiation, which are key aspects of multi-agent
environments.\\ On the other hand, MuZero offers a more flexible framework that
can be adapted to multi-agent environments. Unlike AlphaGo and AlphaGo Zero,
MuZero operates by learning the dynamics of the environment through its
interactions, rather than relying on a fixed model of the world. This approach
allows MuZero to adapt to various types of environments, whether single-agent
or multi-agent, by learning to predict the consequences of actions without
needing explicit knowledge of the environment’s rules. The key advantage of
MuZero in multi-agent settings is its ability to plan and make decisions
without needing to model the entire system upfront. In multi-agent
environments, this ability becomes essential, as MuZero can dynamically adjust
its strategy based on the observed behavior of other agents. By learning not
just the immediate outcomes but also the strategic implications of others'
actions, MuZero can navigate both competitive and cooperative settings.
\begin{figure*}[t]
    \centering
    \includegraphics[width=0.7\textwidth]{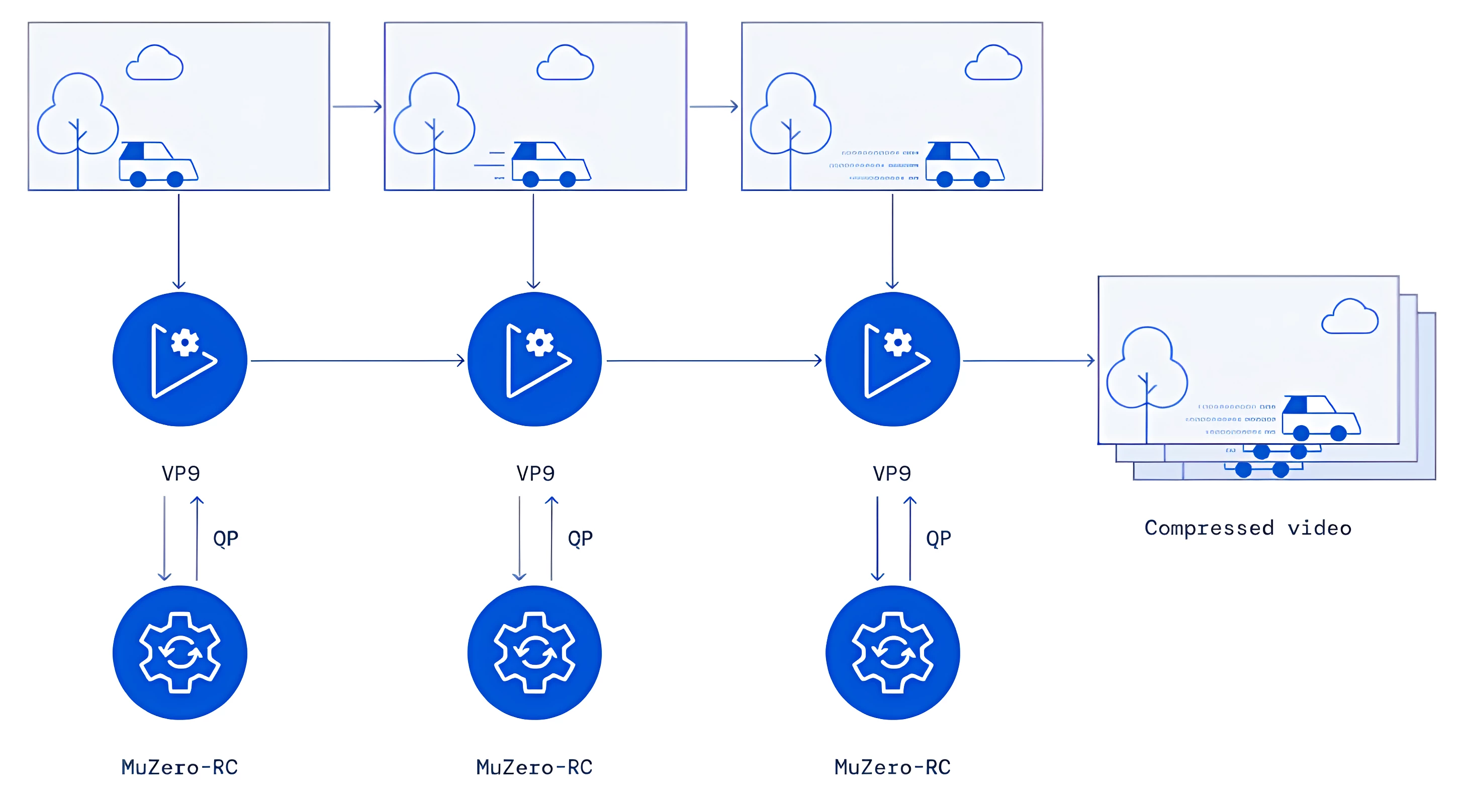}
    \caption{MuZero Rate-Controller (MuZero-RC) optimizing the encoding process in video streaming.}
\end{figure*}

\section{Future Directions}
As mentioned earlier in the paper, The development of the AI models and systems
in the field of gaming represent a good training set for the models to study
the environment, address the challenges, modify the models, and achieve good
results in this field, to judge whether this model is able to be implemented in
real world, and how it can be implemented. The main purpose from such models,
Google DeepMind, through the previous years, had been training the models to
play games, and the main goal was to implement the models of reinforcement
learning in real life, and benefit from them. DeepMind already started in this
implementation with MuZero, and developing other models to be able to be
implemented in real life directly.
\subsection*{MuZero's first step from research into the real world}
One of the notable implementations of MuZero has been in collaboration with
YouTube, where it was used to optimize video compression within the open-source
VP9 codec. This involved adapting MuZero's ability to predict and plan, which
it had previously demonstrated in games, to a complex and practical task of
video streaming. By optimizing the encoding process, as shown in fig. 3, MuZero
achieved an average bitrate reduction of 4\% without degrading video
quality\cite{FD1}. This improvement directly impacts the efficiency of video
streaming services such as YouTube, Twitch, and Google Meet, leading to faster
loading times and reduced data usage for users. This implementation is called
MuZero Rate-Controller (MuZero-RC). Beyond video compression, this initial
application of MuZero outside of game research settings exemplifies how
reinforcement learning agents can address practical real-world challenges. By
designing agents with new capabilities to enhance products across different
sectors, computer systems can be more efficient, less resource-intensive, and
increasingly automated\cite{FD1}.
\subsection*{AlphaFold}
AlphaFold is a model developed by DeepMind that addresses one of the
challenging problems in biology, which is predicting the three-dimensional
structures of proteins from their amino acid sequences. AlphaFold employs
advanced deep learning techniques, prominently featuring reinforcement
learning, to enhance its predictive capabilities.The model operates on a
feedback loop where it generates predictions about protein structures and
receives rewards based on the accuracy of these predictions compared to
experimentally determined structures. This process allows AlphaFold to
iteratively refine its models, optimizing them to better reflect the
complexities of protein folding dynamics. The architecture of AlphaFold
includes deep neural networks that analyze both the sequential and spatial
relationships between amino acids, enabling it to capture intricate patterns
for protein conformation. By training on extensive datasets of known protein
structures, AlphaFold has achieved unprecedented accuracy, often rivaling
experimental methods such as X-ray crystallography and cryo-electron
microscopy\cite{FD2}.\\\\ As shown form the previous models, how the employment
of reinforcementl learning changed starting from making AI systems which play
atari and strategy-based games, reaching to help in human biology and create
protein structures, the enployment of reinforcement learning in games still has
a long journey to be developed which helps in both real life and gaming. Google
DeepMind is still working on other models which are able to be implemented in
real life applications. They also developed models which use the multi-agent
models in games, like AlphaStar, which is a model to play StarCraft II; but
still didn't apply them in real life applications, which is a good future
direction to be developed.

\section*{Conclusion}
Games as an environment for Reinforcement learning, have proven to be very
helpful as a sandbox. Their modular nature enables experimentation for
different scenarios from the deterministic board games to visually complex and
endless atari games. Google's DeepMind utilized this in developing and
enhancing their models starting with AlphaGo that required human gameplay as
well as knowledge of the game rules. Incrementally, they started stripping down
game specific data and generalizing the models. AlphaGoZero removed the need
for human gameplay and AlphaZero generalized the approach to multiple board
games. Subsequently, MuZero removed any knowledge requirements of games and was
able to achieve break-through results in tens of games surpassing all previous
models. These advancements were translated to real-life applications seen in
MuZero's optimization of the YouTube compression algorithm, which was already
highly optimized using traditional techniques. The well defined nature of the
problem helped in achieving this result. Also, AlphaFold used reinforcement
learning in combination with supervised learning and biology insights to
simulate protein structures. While these uses are impressive, especially
coming from models primarily trained to play simple games, they are still
limited in scope. There are many possible holdbacks mainly the training cost,
scalability, and stochastic environments. These models are very expensive to
train despite the limited action and state spaces. This cost would only
increase at more complex environments, taking us to the second issue:
scalability. In many real applications, the actions aren’t mutually exclusive.
This would make the MCTS exponentially more expensive and would further
increase the training cost. Finally, while these models have been tested in
deterministic environments, stochastic scenarios might cause trouble for their
training and inference.

\vspace{12pt}


\begin{thebibliography}{00}
    \bibitem{I1}  N. Y. Georgios and T. Julian, Artificial Intelligence and Games. New York: Springer, 2018.
    \bibitem{I2}  N. Justesen, P. Bontrager, J. Togelius, S. Risi, (2019). Deep learning for video game playing. arXiv.
    \bibitem{I3}  V. Mnih, K. Kavukcuoglu, D. Silver, A. Graves, I. Antonoglou, D. Wierstra, M. Riedmiller, (2013). Playing Atari with deep reinforcement learning. arXiv.
    \bibitem{I4}  A. Graves, G. Wayne, I. Danihelka, (2014). Neural Turing Machines. arXiv.
    \bibitem{I5}  C.J.C.H.Watkins, P. Dayan, Q-learning. Mach Learn 8, 279–292 (1992).
    \bibitem{I6}  DeepMind, (2015, February 12), Deep reinforcement learning.
    \bibitem{I7}  T. Schaul, J. Quan, I. Antonoglou, D. Silver, (2015). Prioritized Experience Replay. arXiv.
    \bibitem{I8}  V. Mnih, A. P. Badia, M. Mirza, A. Graves, T. Lillicrap, T. Harley, D. Silver, K. Kavukcuoglu, (2016). Asynchronous Methods for Deep Reinforcement Learning. arXiv.
    \bibitem{I9}  A. Kailash, P. D. Marc, B. Miles, and A. B. Anil, (2017). Deep Reinforcement Learning: A Brief Survey. IEEE Signal Processing Magazine, vol. 34, pp. 26–38, 2017. arXiv.
    \bibitem{I10} D. Zhao,  K. Shao, Y. Zhu, D. Li, Y. Chen, H. Wang, D. Liu, T. Zhou, and C. Wang, “Review of deep reinforcement learning and discussions on the development of computer Go,” Control Theory and Applications, vol. 33, no. 6, pp. 701–717, 2016 arXiv.
    \bibitem{I11} Z. Tang, K. Shao, D. Zhao, and Y. Zhu, “Recent progress of deep reinforcement learning: from AlphaGo to AlphaGo Zero,” Control Theory and Applications, vol. 34, no. 12, pp. 1529–1546, 2017.
    \bibitem{I12} K. Shao, Z. Tang, Y. Zhu, N. Li, D. Zhao, (2019). A survey of deep reinforcement learning in video games. arXiv.
    
    \bibitem{bg1} L. Thorndike and D. Bruce, Animal Intelligence. Routledge, 2017.
    \bibitem{bg2} R. S. Sutton and A. Barto, Reinforcement learning : an introduction. Cambridge, Ma ; London: The Mit Press, 2018.
    \bibitem{bg3} A. Kumar Shakya, G. Pillai, and S. Chakrabarty, “Reinforcement Learning Algorithms: A brief survey,” Expert Systems with Applications, vol. 231, p. 120495, May 2023
    \bibitem{bg4} Mnih, Volodymyr, et al. “Human-Level Control through Deep Reinforcement Learning.” Nature, vol. 518, no. 7540, Feb. 2015, pp. 529–533.
    \bibitem{Silver2016} D. Silver et al., “Mastering the game of Go with deep neural networks and tree search,” \textit{Nature}, vol. 529, no. 7587, pp. 484–489, Jan. 2016, doi: https://doi.org/10.1038/nature16961.
    \bibitem{Silver2016} D. Silver et al., “Mastering the game of Go with deep neural networks and tree search,” \textit{Nature}, vol. 529, no. 7587, pp. 484–489, Jan. 2016, doi: https://doi.org/10.1038/nature16961.
    \bibitem{agz1} Silver, David, Julian Schrittwieser, Karen Simonyan, Ioannis Antonoglou, Aja Huang, Arthur Guez, Thomas Hubert, et al. 2017. “Mastering the Game of Go without Human Knowledge.” Nature 550 (7676): 354–59. https://doi.org/10.1038/nature24270.

    \bibitem{mz1} J. Schrittwieser et al., “Mastering Atari, go, chess and shogi by planning with a learned model,” Nature, vol. 588, no. 7839, pp. 604–609, Dec. 2020. doi:10.1038/s41586-020-03051-4 
    \bibitem{AD3} DeepMind, "AlphaZero: Shedding New Light on Chess, Shogi, and Go," DeepMind, 06-Dec-2018.
    \bibitem{AD1} T.-R. Wu, H. Guei, P.-C. Peng, P.-W. Huang, T. H. Wei, C.-C. Shih, Y.-J. Tsai, (2023). MiniZero: Comparative analysis of AlphaZero and MuZero on Go, Othello, and Atari games. arXiv.
    \bibitem{AD2} K. Zhang, Z. Yang, T. Ba\c{s}ar, (2021). Multi-agent reinforcement learning: A selective overview of theories and algorithms. arXiv preprint arXiv:2103.04994

    
    \bibitem{FD1} "MuZero’s first step from research into the real world," DeepMind, Feb. 11, 2022.
    \bibitem{FD2} Jumper, J et al. Highly accurate protein structure prediction with AlphaFold. Nature (2021)

\end{thebibliography}
\end{document}